\documentclass{article}
\usepackage{ijcai22}

\usepackage{times}
\usepackage{soul}
\usepackage{url}
\usepackage[hidelinks]{hyperref}
\usepackage[utf8]{inputenc}
\usepackage[small]{caption}
\usepackage{graphicx}
\usepackage{amsmath}
\usepackage{amsthm}
\usepackage{booktabs}
\usepackage{algorithm}
\usepackage{algorithmic}
\usepackage{bm}
\usepackage{amssymb}
\usepackage{xcolor}
\usepackage{makecell}
\usepackage{colortbl}
\usepackage{textcomp}
\definecolor{maroon}{cmyk}{0,0.87,0.68,0.32}

\usepackage[capitalise]{cleveref}

\newcommand{\bh}{\mathbf{h}}
\newcommand{\bs}{\mathbf{s}}
\newcommand{\bp}{\mathbf{p}}

\newcommand{\bA}{\mathbf{A}}

\newcommand\Tstrut{\rule{0pt}{2.2ex}}       % "top" strut
\newcommand\Bstrut{\rule[-0.9ex]{0pt}{0pt}} % "bottom" strut
\newcommand{\note}[2]{\textcolor{#1}{#2}}

\newcommand{\mred}[1]{\note{black}{#1}}

\newcommand*\samethanks[1][\value{footnote}]{\footnotemark[#1]}
\graphicspath{{./}{pics/}}

\title{``Think Before You Speak'': \mred{Improving Multi-Action Dialog Policy by Planning Single-Action Dialogs}}

\author{
Shuo Zhang$^1$
\and
Junzhou Zhao$^1$\thanks{ Corresponding Author }\and
Pinghui Wang$^1$\samethanks[1]\and
Yu Li$^1$\and
Yi Huang$^2$ \and
Junlan Feng$^2$ \\
\affiliations
$^1$MOE KLINNS Lab, Xi’an Jiaotong University, Xi’an 710049, P. R. China\\
$^2$JIUTIAN Team, China Mobile Research, Beijing 100053, P. R. China\\
\emails
\{zs412082986, liyu1998\}@stu.xjtu.edu.cn,
\{junzhou.zhao, phwang\}@mail.xjtu.edu.cn,\\
\{huangyi, fengjunlan\}@chinamobile.com
}

\begin{document}

\maketitle

\begin{abstract}
  Multi-action dialog policy (MADP), which generates multiple atomic dialog actions per
turn, has been widely applied in task-oriented dialog systems to provide
expressive and efficient system responses.
Existing MADP models usually imitate action combinations from the labeled
multi-action dialog samples.
Due to data limitations, they generalize poorly toward unseen dialog flows.
\mred{While interactive learning and reinforcement learning algorithms can be applied to
incorporate external data sources of real users and user simulators,}
% While robustness training algorithms (e.g., reinforcement learning) are applied to
% incorporate external data sources (e.g., user simulator) for model training, 
they take significant manual effort to build and suffer from instability.
To address these issues, we propose Planning Enhanced Dialog Policy (PEDP), a
novel multi-task learning framework that learns single-action dialog dynamics to enhance multi-action prediction.
% utilizes single-action dialog samples to enhance multi-action prediction.
Our PEDP method simulates single-action dialog fragments with model-based planning to conceive what to express before deciding the current response.
% It simulates a single-action dialog procedure through model-based planning to
% discovers contextually relevant contents before multi-action prediction through simulating a single-action dialog.
%  conceivingwhat to express before deciding the current response 
Experimental results on the MultiWOZ dataset demonstrate that our fully supervised learning-based method achieves a solid task success rate of $90.6\%$, improving $3\%$ compared to the state-of-the-art methods\footnote{The source code and the appendix of this paper can be obtained from \url{https://github.com/ShuoZhangXJTU/PEDP}.}.

% Multi-action dialog policy, which generates multiple atomic actions per turn,
% has been widely applied in task-oriented dialog systems to provide more
% expressive and efficient system responses.
% However, while existing supervised learning based methods share the flaw of low
% robustness to unseen real-world human-machine dialog flows, robustness
% training-based methods (e.g., reinforcement learning) take too many manual
% efforts and suffer from instability.
% % Moreover, these issues are magnified in the multi-action setting on account of
% % numerous and complex action combination patterns.
% In this work, we propose a novel {\em Planning Enhanced Dialog Policy} (PEDP)
% framework that achieves robust and explainable multi-action dialog policy with
% solely supervised learning.
% Unlike seeking for external data source, our PEDP model further utilizes
% single-action dialog samples to enhance multi-action prediction.
% It simulates a single-action dialog procedure that looks ahead several turns
% to discover contextually relevant contents before multi-action prediction.
% % Following the human dialog strategy of ``think before you speak'', our PEDP
% % model plans several single-action paths before ensemble multi-action
% % prediction to effectively discover and condense contextually relevant actions.
% Experiments conducted on the MultiWOZ dataset show the promising performance of
% our method with a marginal improvement even comparing to the state-of-the-art
% robustly trained approaches.

%%% Local Variables:
%%% mode: latex
%%% TeX-master: "main"
%%% End:

 \end{abstract}

\section{Introduction}

{\em Task-oriented Dialog Systems} (TDS) assist users in completing specific
tasks (e.g., restaurant booking) using natural language, and have been applied to
various commercial
services~\cite{zhang2020recent,qin2021suma,zhang2020contract}.
In a TDS, the dialog policy, which decides the next system actions according to the
current dialog context, plays a vital role as it dramatically influences the
dialog efficiency (e.g., the conciseness and smoothness).

\begin{figure}[ht]
	\includegraphics[width=.48\textwidth]{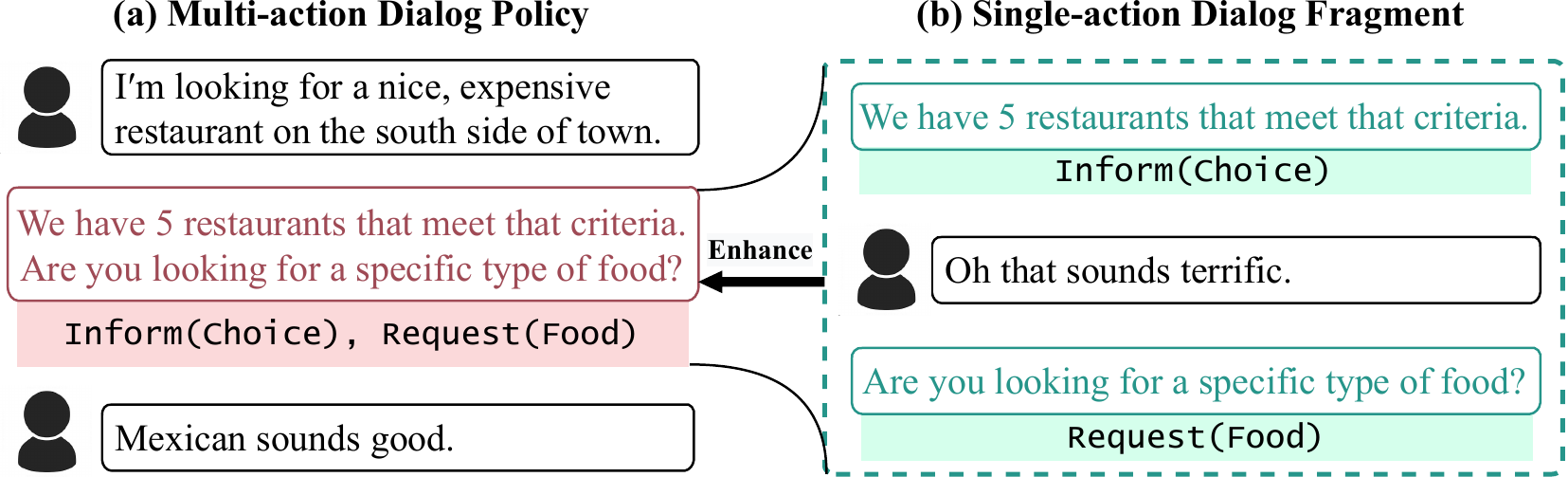}
	\centering
	\caption{(a) Example dialog under multi-action dialog policy\protect\footnotemark. We propose to learn single-action dialog dynamics (b) to model conditional act combination patterns and enhance multi-action prediction.}
	\label{fig:examplecase}
\end{figure}

Prior work on dialog policy typically assumes that the agent only produces one
action per turn~\cite{mnih2015human,peng2018deep}, which leads to lengthy
dialogs and increases the risk of task failure.
To address these weaknesses, some recent works have investigated {\em Multi-Action
  Dialog Policy Learning} (MADPL) that generates multiple actions concurrently as
the current system response to boost the expressive power and efficiency of the
TDS~\cite{shu2019modeling,jhunjhunwala2020multi}.
MADPL can be formulated as a multi-label classification or sequence generation
problem, and thus solved by {\em Supervised Learning} (SL) based methods that
imitate action combinations from labeled dialog
samples~\cite{shu2019modeling}.

Despite their success, existing SL-based methods often
\mred{generalize poorly towards}
% suffer from poor robustness for
real-world human-machine dialog flows~\cite{jhunjhunwala2020multi}.
% ~\cite{liu2018dialogue,jhunjhunwala2020multi}.
% A key challenge for MADPL is to achieve robustness towards real-world
% human-machine dialog flows with limited supervision costs (i.e., data
% annotation costs, deployment costs, etc.).
Task-oriented dialogs are known to share the ``one-to-many'' nature that various
appropriate responses may exist under the same
context~\cite{huang2020semi}.
Accordingly, in the multi-action setting, various action combinations could be
considered appropriate under the same context, which are mostly not covered in
limited human-labeled dialog datasets~\cite{jhunjhunwala2020multi}.
With straightforward imitation, existing SL-based methods will be restricted to a
small subspace of the entire action space, leading to limited generalization ability towards unseen dialog scenarios.
% towards scenarios never seen during training.

\footnotetext{A dialog policy responses by predicting atomic dialog actions
    represented as ``Domain-Intent(Slot)'' phrases.
    We omit the domain (``restaurant'') for clarity.}

% The model need to model highly complex action co-occurrence patterns.
% Cost-prohibitive human-labeled dialog datasets are usually limited in size
% that is unlikely to cover a large number of possible act combinations.

% MADPL is first addressed with {\em Supervised Learning} (SL) based methods that
% imitate action combinations demonstrated in human-human dialog
% samples~\cite{shu2019modeling}.
% However, cost-prohibitive human-labeled dialog datasets are usually limited in
% size that is unlikely to cover a large number of possible act combinations.
% With straight-forward imitation, existing SL-based methods will be restricted to
% a small subspace of the complete output action space, which leads to poor
% robustness towards real-world human-machine dialog flows and lowers service
% quality.

% \llnote{Some existing studies try to address the above issue from three
% different perspectives: (1) data augmentation...
% (2) reinforcement learning...
% (3) adversarial learning...
% However, ...}
Some existing studies try to address the above issue from three different
perspectives: 1) {\em Interactive Learning}~\cite{jhunjhunwala2020multi}, which
performs data augmentation through human-machine interactions; 2) {\em
  Reinforcement Learning} (RL)~\cite{takanobu2019guided,zhu2020convlab}, which
enables learning from online human-interacting dialogs; and 3) {\em Adversarial
  Learning} (AL)~\cite{takanobu2019guided,li2020rethinking}, which provides extra
supervision signals with a machine-learned discriminator.
However, the performance gain is obtained at the cost of additional supervision.
% (i.e., data annotation costs, deployment costs, etc.).
For example, they require tons of human work to label dialogs, build real-world
environments such as user simulators, and design learning strategies.
Moreover, RL and AL-based algorithms suffer from the unstable training process~\cite{2020Stable,tian2020deep},
bringing extra deployment costs in real-world applications.

In this work, we present {\em Planning Enhanced Dialog Policy} (PEDP), an SL-based multi-task learning framework for improved MADPL.
\mred{Unlike prior methods that directly imitate complex action combination patterns from limited multi-action dialog \textit{\textbf{turns}}, we propose to further learn such patterns from single-action dialog \textit{\textbf{fragments}}.}
We find that one turn in the multi-action dialog usually corresponds to several turns in the single-action setting (i.e., fragments) to achieve the similar (or identical) dialog state transition (see \cref{fig:examplecase}).
\mred{This phenomenon inspires us to learn to simulate single-action dialog fragments for enhancing multi-action prediction through discovering and incorporating contextually relevant dialog actions.}
% This phenomenon inspires us to discover and incorporate contextually related content through single-action dialog dynamics before multi-action prediction.
% dialog actions to enhance multi-action prediction.
To materialize this idea, 
% we propose to leverage single-action dynamics to
% discover and incorporate contextually related dialog actions before multi-action prediction.
we integrate model-based planning into the decision process and propose a {\em Single-action Dialog Planning} module.
\mred{It plans several single-action paths (dialog fragments) based on the current dialog state before multi-action prediction. Each path looks ahead several turns through ``self-play'' between a discrete dialog policy model and a world model that imitates user behaviors. We further propose an {\em Ensemble Prediction} module that aggregates the multi-action prediction probabilities from multiple paths to reduce the impact of low-quality dialogs.}
% \mred{The planning result serves as addtional , which contains the discovered contextually relevant dialog actions, is used to aid in predicting the dialog actions that should be executed concurrently.}
% Specifically, a world model that simulates user behaviors 
% that simulate a single-action dialog procedure.
% looks ahead several dialog turns to simulate a  through cyclical interaction between a discrete dialog policy model and a
% world model that simulates user behaviors.
% under single-action dynamics We observe that a macro-action\footnote{A
% macro-action is considered a set of actions (without sequential dependency)} is
% usually needed when interacting with the corresponding single actions results in
% a similar (or identical) dialog state transition but leads to redundant dialog
% turns (see Figure~\ref{fig:examplecase}).
% Accordingly, we may view the execution of multi-action as a single-action
% dialog procedure and investigate single-action dialog samples for MADPL.
% To this end, we propose to look ahead several dialog turns under single-action
% dynamics ({\em Planning}) to discover and incorporate contextually related
% dialog actions before multi-action prediction.
% Specifically, the planning is performed through cyclical interaction between a
% discrete dialog policy model and a model of the environment (i.e., world model)
% that updates the current state given the atomic action.
% And the macro-action is predicted by sampling from the probabilities that are
% aggregated from each path.
Unlike seeking external data sources, our method better uses existing dialog samples.
It performs multi-action prediction in a reasonable and explainable way, thus boosting performance towards unseen dialog flows.
% corpus with single-action dialog samples.
% joint imitation of single-action 
% multi-task learning of single-act prediction, dialog state transition, and multi-action prediction. 
% and incorporate single-action dialog samples to model complex action
% co-occurrence patterns.
% With multi-task modeling of the environment (i.e., state transition),
% single-action policy, and multi-action policy.

% The proposed PEDP model is jointly trained with multi-tasks of state recovery,
% single-action prediction, stop flag prediction, and macro-action generation to
% fully leverage the training corpus.
% The proposed PEDP model is concisely composed of RNNs and Feed-forward Neural
% Networks for planning and rebuilding respectively and is jointly trained with
% multi-tasks of state prediction, single-action prediction, and macro-action
% generation.

To evaluate the proposed method, we conduct experiments on the MultiWOZ~\cite{budzianowski2018multiwoz} and the SGD~\cite{rastogi2020towards} datasets.
The experimental results demonstrate that the extension of planning
brings significant improvement to our model's performance and dialog efficiency,
resulting in a promising task success rate of $90.6\%$, improving $3\%$ compared to the state-of-the-art methods. 

% and then condense the anticipatory actions into a macro-action for simplicity.
% Considering only the dialog procedure over single-action dynamics (see top of
% Figure~\ref{fig:examplecase}), we first look ahead a few dialog turns to find
% consecutive single-actions to execute ({\em Planning}) and then condense the
% anticipatory actions into a macro-action\footnote{A macro-action is considered a
% set of actions (without sequencial dependency)} for simplicity ({\em
% Rebuilding}).

%%% Local Variables:
%%% mode: latex
%%% TeX-master: "main"
%%% End:

\section{Related Work}
\noindent
\textbf{Multi-Action Dialog Policy Learning (MADPL).}
Recent efforts have been made to learn a multi-action dialog policy for task-oriented dialogs where the agent can produce multiple actions per turn to expand its expressive power.
\mred{Some attempts~\cite{zhu2020convlab,gordon2020learning} (primarily DQN-based methods)} incorporate the top-$k$ most frequent dialog action combinations in the dataset into the output space and \mred{simplify} MADPL as a multi-class classification problem.
However, the expressive power \mred{and flexibility} of the dialog agent are severely limited. 
Considering the sequential dependency among the acts, \citeauthor{shu2019modeling}~(\citeyear{shu2019modeling})
proposes to formulate MADPL as a sequence generation problem.
\mred{However, 
the order in which people express their intentions is complicated and has not been properly annotated in the existing human-to-human dialog datasets\footnote{The sequential order of the actions in the annotated samples is usually inconsistent with the expressive order in text responses.}.}
The discrepancy between the task setting and data annotation can lead to misleading performance.
% However, in human-human dialogs, the order in which people express their
% intentions is highly complex or even random since it is conditioned on both the
% context and personal habits.
% For example, it is okay to inform a hotel's price and address in any sequential
% order, while it is usually preferred to inform the food's price before further
% asking the payment method.
% To our best knowledge, such sequential order has not been properly
% considered and correctly labeled in existing datasets, which brings a practical challenge of data limitation.
% \cite{shu2019modeling} applies a fixed order (e.g., alphabetical order).
Recent works~\cite{shu2019modeling,li2020rethinking} typically cast MADPL as a multi-label classification problem and address it with supervised learning.
However, the complex action combinations can
exponentially enlarge the output space~\cite{jhunjhunwala2020multi}.
The limited coverage of the output action space of the existing dialog corpus will greatly hinder SL-based methods' performance. 
\citeauthor{jhunjhunwala2020multi}~(\citeyear{jhunjhunwala2020multi}) addresses this problem by data augmentation through interactive learning.
The problem can also be partially alleviated through RL-based~\cite{zhu2020convlab} and AL-based
methods~\cite{takanobu2019guided,li2020rethinking} by allowing the model to explore more potential dialog scenarios.
Unlike existing works that ``memorize'' action combinations from limited multi-action samples, we mimic the human decision process and discover dialog actions to express before condensing them 
into a brief response through simulating single-action dialogs.

\vskip 6pt
\noindent
\textbf{Planning in Dialog Policy Learning.}
Planning refers to the process that uses a model of the environment to improve a
policy.
% Planning based RL methods have been introduced for task-completion dialog
% policy learning to improve data efficiency and boost overall performance.
\citeauthor{peng2018deep}~(\citeyear{peng2018deep}) first introduces background planning for task-completion
dialog policy learning.
They propose to use Deep Dynamic Q-network, in which a world model is trained to
mimic real-world users and provide extra simulated experience for training.
The integration of planning can substantially improve data efficiency.
Since the world model is not perfect and may generate erroneous
experiences (state-action pairs) that hinder policy learning.
\citeauthor{su2018discriminative}~(\citeyear{su2018discriminative}) proposes to train a discriminator to filter out
low-quality simulated experiences.
\citeauthor{wu2019switch}~(\citeyear{wu2019switch}) design a switcher-based mechanism to balance
the use of real and simulated experiences automatically.
\citeauthor{wang2020task}~(\citeyear{wang2020task}) proposes decision-time planning that adopts a rollout
algorithm (Monte-Carlo Tree Search) to boost the accuracy of action value
estimations.
In summary, previous work has focused on RL-based single-action dialog policy
learning.
Instead, we address SL-based multi-action dialog policy learning and extend
planning to generate anticipatory single-action dialog trajectories for further
combination.

%%% Local Variables:
%%% mode: latex
%%% TeX-master: "main"
%%% End:

% \input{s_task}

\begin{figure*}[htp]
  \includegraphics[width=.9\textwidth]{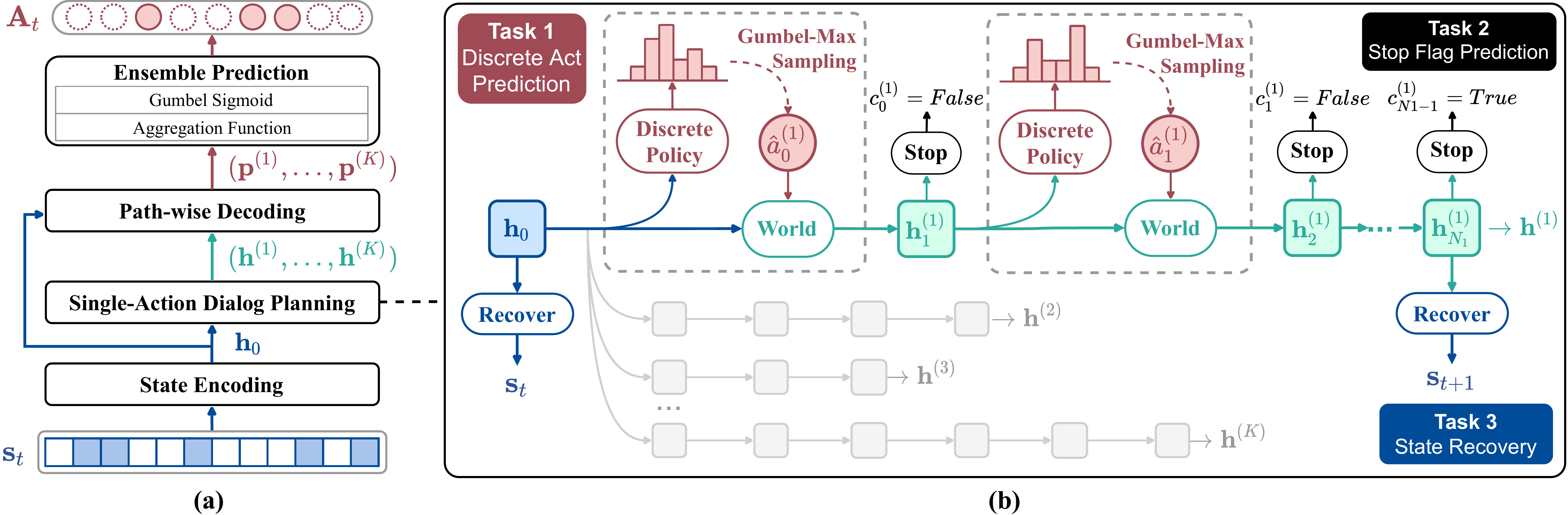}
  \centering
  \caption{(a) The Planning-Enhanced Dialog Policy (PEDP) framework.
    It utilizes a {\em single action dialog planning} module (b) to incorporate contextually relevant contents before multi-action prediction. 
    A total of $K$ single-action dialog procedures are planned, with the $k$-th path looking ahead $N_k$ steps
    under single-action dialog dynamics.
    At each step, the discrete policy model predicts an atomic dialog action
    $a_{n}$ given the previous dialog state embedding $\bh_{n-1}$.
    The world model, which simulates user behavior, responds to the predicted action $a_{n}$ and updates the dialog state embedding from $\bh_{n-1}$ to $\bh_{n}$.}
  \label{fig:framework}
\end{figure*}

\section{Methodology}
In this section, we introduce the {\em Planning-Enhanced Dialog Policy} (PEDP)
framework, an SL-based multi-task learning framework for improved MADPL.

\subsection{Task Formulation}
We regard MADPL as a multi-label classification task, i.e., given the current dialog
state that summarizes the dialog history, we want to predict a {\em macro-action} that is a set of
% that includes several sequentially
independent {\em atomic dialog actions} that serves as the current system response.
Following~\cite{li2020rethinking}, each atomic dialog action is a concatenation of domain name, action type, and
slot name, e.g., ``hotel-inform-area''.
% \footnote{There are 166 atomic dialog actions in total in the action space.
% }.

\subsection{Overview}

The PEDP framework is inspired by the human dialog strategy of ``think before
you speak'' that conceives relevant content to express before deciding the current
response.
\mred{We first discover contextually relevant dialog actions by looking ahead several turns under the single-action dialog dynamics (i.e., {\em Single Action Dialog Planning})} and then use the
planning result as guidance to predict the atomic dialog actions that should
be executed concurrently (i.e., macro-action).

Formally, let $\bs_t$ denote the current dialog state representing the
dialog history up to the current dialog turn.
We encode $\bs_t$ into a dialog state embedding $\bh_t$.
Given the current state embedding $\bh_t$, we plan $K$ independent single-action
dialog paths, and the $k$-th dialog path is represented by a vector
$\bh^{(k)}$, $k=1,\ldots, K$.
Our model then decodes each dialog path to a probability distribution over
atomic dialog actions, i.e., $\bp^{(k)}$.
Finally, these distributions $\{\bp^{(k)}\}_{k=1}^K$ are aggregated to form a
unified distribution, from which atomic dialog actions in the macro-action
$\bA_t$ are sampled.

Figure~\ref{fig:framework} depicts the architecture of PEDP, which mainly consists
of four modules: 1) {\em State Encoding}, which encodes the dialog state into a
compact vector; 2) {\em Single-Action Dialog Planning}, which plans several
independent dialog paths with each one looking ahead several steps considering
only the single-action dialog dynamics; 3) {\em Path-wise Decoding}, which
predicts the multi-action probability distribution for each dialog path; and 4)
{\em Ensemble Prediction}, which aggregates the probabilities and samples the
final macro-action.
In what follows, we elaborate on each of these modules in PEDP.

\subsection{State Encoding}

The dialog state is a summary of the dialog history.
Inspired by~\cite{li2020rethinking}, we propose to include the following
information into dialog state $\bs_t$ at time step $t$: 1) embedding of returned
entities for a query, 2) the last user action, 3) the last system action, and 4)
the belief state that contains the requested slots and informed slots of user and
agent.
% The final current-turn dialog state $\bs_t$ is a vector of $553$ bits.

To densify the dialog state, we apply a Fully-connected Feed-forward
Network (FFN) to obtain a compact dialog state embedding $\bh_t$.
The FFN consists of two linear transformations with a ReLU activation in the
middle layer, i.e.,
\[
  \bh_t = \texttt{FFN}_{enc}(\bs_t) = \texttt{ReLU}(\bs_tW_1 + b_1)W_2 + b_2.
\]
In what follows, this dialog state embedding $\bh_t$ will serve as the initial
dialog state embedding for planning.

\subsection{Single-Action Dialog Planning}

As discussed in the introduction, the straight-forward multi-label imitation of action
co-occurrence suffers from the poor generalization ability issue due to the lack of labeled
multi-action training data.
To address this issue, we propose a {\em single-action dialog planning} module that further leverages single-action dialog samples to model contextual action co-occurrence patterns.
It looks ahead several steps under single-action dialog dynamics to predict the
next dialog state, which will serve as the additional information to enhance
macro-action prediction.
In what follows, we describe how to model the dialog state transition in a
single-action dialog.

Given the initial dialog state embedding $\bh_t$ at the current dialog turn $t$, we plan $K$ independent
single-action dialogs to discover contextually related atomic dialog actions.
In each dialog, a {\em discrete policy model} and a {\em world model} interact to look ahead several steps.
Let $\bh_{t,n}^{(k)}$ denote the dialog state embedding at the $n$-th step in the
$k$-th dialog for $n=0,\ldots,N_k$ where $N_k$ is the length of the $k$-th dialog,
and $\bh_{t,0}^{(k)}=\bh_t$, $\bh_{t,N_k}^{(k)}=\bh^{(k)}$.
The last dialog state embedding $\bh^{(k)}$ estimates the hidden vector of the future dialog state
  $\bs_{t+1}$ and summarizes the planned single-action dialog.
In what follows, we describe how to plan a single step from $\bh^{(k)}_{t,n}$ to
obtain $\bh^{(k)}_{t,n+1}$.
To simplify the notations, we omit the dialog index $k$ and dialog turn $t$.

Given the dialog state embedding $\bh_n$ at step $n$, we want first to predict a
single action $a_n$ and then use it to update $\bh_n$ to obtain $\bh_{n+1}$.
This can be achieved by cyclical interactions between a discrete policy model (DP)
that predicts an atomic dialog action under the current-step dialog state and a world
model that simulates user behavior to update the current-step dialog state, i.e.,
\begin{gather*}
  a_n =\texttt{DP}(\bh_n)
        \triangleq\mathrm{GumbelSoftmax}^{(\tau_d)}(\bh_nW_d+b_d) \\
  \bh_{n+1} =\texttt{World}(\bh_n, a_n)
             \triangleq\texttt{GRU}(\bh_n, \texttt{Emb}(a_n)).
\end{gather*}
Here, DP is implemented as a single linear layer followed by a Gumbel-Softmax
function~\cite{jang2016categorical} parameterized by $\tau_d$.
The Gumbel-Softmax function draws an atomic dialog action sample from a categorical
distribution, diversifying the planned dialogs.
$\tau_d$ is selected to balance the approximation bias and the magnitude of
gradient variance.
The world model is implemented using a GRU to
model dialog state transitions, and $\texttt{Emb}(a_n)$ denotes the embedding
vector of atomic dialog action $a_n$.

We observe that both executing a macro-action and sequentially executing the corresponding atomic dialog actions will lead to a similar (or identical) state
transition (see \cref{fig:examplecase}).
Accordingly, we assume that the planned information is adequate once such a state
transition is achieved, and propose to determine if to stop planning by comparing
the initial and the \textbf{next} dialog state embeddings, i.e., $\bh_0$ and $\bh_{n+1}$ (see \cref{fig:framework}b).
This is also modeled using a neural network, i.e.,
\[
  c_n = \mathrm{GumbelSoftmax}^{(\tau_s)}(\texttt{FFN}_{st}([\bh_0:\bh_{n+1}]))
\]
where $c_n$ is a binary variable, ``$:$" denotes vector concatenation, and
$\texttt{FFN}$ is a 2-layer fully-connected feed-forward network using the ReLU
activation function in the middle layer.

Note that the single-action dialog planning module estimates dialog state
transitions in the dialog state embedding space from $\bh_0$ to $\bh_{N}$, and it
demands supervision from the labeled structural dialog state transition samples
from $\bs_t$ to $\bs_{t+1}$.
To enable model training, we incorporate a recovery model for mapping the state embeddings back to the structured state-space.
Formally, we have,
\begin{gather*}
  \bs_t = \texttt{Recover}(\bh_0) \\
  \bs_{t+1} = \texttt{Recover}(\bh_N)
\end{gather*}
Here, $\texttt{Recover}$ is implemented by a 2-layer FFN 
and is only used during the training stage.

\subsection{Path-wise Decoding}
Leveraging the planned single-action dialog, we predict the probability distribution over atomic dialog actions.

Given the initial and the last dialog state embeddings $\bh_0$ and $\bh^{(k)}$, we decode the multi-action probability distribution $\mathbf{p}^{(k)}$ with a neural network applied to each path separately and identically.
Since our task formulation posits no dependency among atomic dialog actions, following~\cite{li2020rethinking}, we perform binary classification for each specific action in the atomic dialog action space independently to decide whether it is selected.
Specifically, we instantiate the decoder 
$
  \mathbf{p}^{(k)} = [\mathbf{p}^{(k)}_1:\ldots:\mathbf{p}^{(k)}_{M}]
$,
where $k$ refers to the planned path and $M$ is the size of the action space.
Each $\mathbf{p}^{(k)}_m, m=1, \ldots, M$ is a vector computed as:
\[
  \mathbf{p}^{(k)}_m = \texttt{FFN}^{dec}_m([\bh_0:\bh^{(k)}]).
\]
% which is 166 in our case. 
Future efforts at
exploring different decoders may further improve the performance.

\subsection{Ensemble Prediction}
The performance of the proposed model is affected by the quality of the planned dialogs.
To reduce the impact of low-quality dialogs, we propose to ensemble the multi-action probability distributions with an aggregation function, i.e.,
\[
  \mathbf{P}_t = \mathrm{Aggr}(\mathbf{p}^{(1)}, \ldots, \mathbf{p}^{(K)})
\]
where $\mathrm{Aggr}(\cdot)$ is the mean average in our case. 

Task-oriented dialogs share a critical ``one-to-many'' nature that different responses may be proper under the same context, making it essential to incorporate stochastic factors to build a multi-action dialog policy.
We empirically verified that this stochasticity could be achieved with probability sampling.
In our case, we apply the Gumbel-Sigmoid function to sample the macro-action, that is,
\[
  \mathbf{A}_t = \mathrm{GumbelSigmoid}(\mathbf{P}_t) = \frac {e^{(\mathbf{P}_t+g_1)/\tau}}{e^{(\mathbf{P}_t+g_1)/\tau} + e^{(\mathbf{P}_t+g_2)/\tau}}
\]
Here $\mathrm{GumbelSigmoid}(\cdot)$ is a modification of the Gumbel-Softmax function, regarding sigmoid as a softmax with two logits $p$ and $0$.
$\tau$ denotes the temperature factor, $g_1$ and $g_2$ are Gumbel noises.

\subsection{Training}

We use a multi-task objective to train all the modules jointly. 

\vskip 6pt
\noindent\textbf{Task 1: Discrete Act Prediction (DAP)}~~
For the planned single-action sequence $\bm{a}=(a_0, \ldots, a_{N-1})$, we aim to
maximize the log-likelihood (MLE) for the joint probability
$p(\boldsymbol{a}|\bh_0)$, which can be factorized as:
\begin{align*}
  p(\boldsymbol{a}|\bh_0)= p_\theta(a_0|\bh_0)\prod_{n=1}^{N-1}{\underbrace{p_{\theta}(a_{n}|\bh_{n})}_\text{DAP}\underbrace{
  p_{\phi}(\bh_{n}|a_{n-1}, \bh_{n-1})}_\text{state transition}}
\end{align*}
where $\theta$ and $\phi$ denotes trainable parameters for the discrete dialog
policy model and the world model, respectively.

\vskip 6pt
\noindent\textbf{Task 2: Stop Flag Prediction (SFP)}~~
We define the objective of predicting the stop flag sequence
$\bm{c}=(c_0, \ldots, c_{N-1})$ as MLE for the joint probability
$p(\boldsymbol{c}|\bh_0)$, which can be factorized as:
\begin{align*}
  p(\boldsymbol{c}|\bh_0) = \prod_{n=0}^{N-1}
                  \underbrace{p_{\gamma}(c_n|\bh_{n+1}, \bh_0)}_\text{SFP}
                  \underbrace{p_{\phi, \theta}(\bh_{n+1}|\bh_{n})}_\text{1-step planning}
\end{align*}
where $\gamma$ parameterizes the stop prediction model, the joint probability of $p_{\phi, \theta}(\bh_{n+1}|\bh_{n})$ is
factorized as $p_{\phi}(\bh_{n+1}|a_{n}, \bh_{n})
p_{\theta}(a_{n}|\bh_{n})$ of state transition and discrete act
prediction.

\vskip 6pt
\noindent\textbf{Task 3: State Recovery (SR)}~~
We consider a state recovery objective to supervise state encoding and state
transition.
Precisely, we predict the current dialog state $\bs_t$ and next dialog
state $\bs_{t+1}$ with the initial dialog state embedding $\bh_0$ and the last state embedding $\bh_N$, respectively, i.e.,
\begin{align*}
  p(\bs_t)&=\underbrace{p_\zeta(\bs_t|\bh_0)}_\text{SR}\underbrace{p_\eta(\bh_0|\bs_t)}_\text{state encoding} \\
  p(\bs_{t+1}|\bs_t)&=\underbrace{p_\zeta(\bs_{t+1}|\bh_N)}_\text{SR}\underbrace{p_\eta(\bh_0|\bs_t)}_\text{state encoding}
                  \prod_{n=0}^{N-1}{\underbrace{p_{\phi, \theta}(\bh_{n+1}|\bh_{n})}_\text{1-step planning}}
\end{align*}
where $\eta$ and $\zeta$ denotes trainable parameters for state encoder and the $\texttt{Recover}$, respectively. The joint probability $p_{\phi, \theta}(\bh_{n+1}|\bh_{n})$ is the same as explained in Task 2.
% sequentially factorized as $\prod^N_1{p_{\phi}(\bh_n|a_{n-1},
%   \bh_{n-1}) p_{\theta}(a_{n-1}|\bh_{n-1})}$ in a recurrent manner.
\vskip 6pt
\noindent\textbf{Task 4: Multi-Action Prediction (MAP)}~~
Finally, we introduce a multi-action prediction objective to supervise all the modules in our PEDP framework. That is:
\begin{align*}
  p(\bA_{t}|\bs_t)&=\underbrace{p_\omega(\bA_{t}|\bh_0,\bh_N)}_\text{MAP}\underbrace{p_\eta(\bh_0|\bs_t)}_\text{state encoding}
                  \prod_{n=0}^{N-1}{\underbrace{p_{\phi, \theta}(\bh_{n+1}|\bh_{n})}_\text{1-step planning}}
\end{align*}
where $\omega$ denotes trainable parameters for the decoder. The rest is the same as explained in Task 3.

\vskip 6pt
\noindent\textbf{Full Objective}~~
The overall loss is the weighted sum of four losses: two cross-entropy losses of DAP and SFP and two binary cross-entropy losses of SR and MAP\footnote{We refer the reader to Appendix A and B for data usage and model implementation details.}.

%%% Local Variables:
%%% mode: latex
%%% TeX-master: "main"
%%% End:

\section{Experiments}
% \begin{table}[tp]
% \centering
% \small
% \begin{tabular}{l|cccccc}
% \hline
% \hline
%     & \textbf{0} & \textbf{1} & \textbf{2} & \textbf{3} & \textbf{4} & \textbf{\textgreater= 5}  \\ \hline
% Train & 2143   & 16161  & 16302  & 9646   & 6001  & 6497   \\ \hline
% Valid  & 138   & 2038  & 2216  & 1306   & 781  & 886 \\ \hline
% Test & 131   & 2050  & 2214  & 1344   & 770  & 863  \\ \hline
% \end{tabular}
% \caption{dialog act counts by turn.}
% \label{tab: data_act_count}
% \end{table}

\subsection{Settings}
\textbf{Evaluation Metrics}~~
We perform the automatic evaluation from two perspectives:

1) \mred{{\em Interactive Evaluation}}, which evaluates task completion quality through dialog interaction with the user simulator.
We use {\em Inform F1}, {\em Inform Recall}, {\em Match}, {\em Turn}, and {\em Success} as evaluation metrics.
{\em Inform F1} and {\em Inform Recall} evaluate whether all the requested information (e.g., hotel address) has been provided.
{\em Match} evaluates whether the booked entities match the indicated constraints (e.g., the cheapest restaurant in town).
{\em Turn} and {\em Success} evaluate the efficiency and level of task completion of dialog agents, respectively.
Specifically, a dialog is considered successful if all the requested information is provided (i.e., {\em Inform Recall}~$= 1$) and all the entities are correctly booked (i.e., {\em Match}~$= 1$).
% {\em Success} is a binary integer for each session.

2) {\em Standard Evaluation}, which measures the generalization ability with the test set of the training corpus. 
We report the sample-wise {\em Precision}, {\em Recall}, and {\em $F_1$} score of the macro-actions, where each atomic dialog action is assumed to be correct if it is included in the ground truth.

% For the robust evaluation, we use {\em Inform F1}, {\em Inform Recall}, {\em Match}, {\em Turn}, and {\em Success} as evaluation metrics.
% {\em Inform F1} and {\em Inform Recall} evaluates whether all the requested information (e.g., hotel address) has been provided, while {\em Match} evaluates whether the booked entities match the indicated constraints (e.g., the cheapest restaurant in town).
% {\em Turn} and {\em Success} evaluate the efficiency and level of task completion of dialog agents, respectively.
% Specifically, a dialog is considered successful only if all the requested information is provided (i.e., {\em Inform Recall}~$= 1$) and all the entities are correctly booked (i.e., {\em Match}~$= 1$).
% % {\em Success} is a binary integer for each session.
% For the standard evaluation, we report the sample-wise {\em Precision}, {\em Recall}, and {\em $F_1$} score of the macro-actions, where each atomic action is consumed to be correct if
% % both the intent and the slot match 
% it is included in the ground truth.
\vskip 6pt
\noindent\textbf{Datasets}~~
% \subsubsection{Datasets}
We conduct experiments on two dialog datasets.

1) \textit{MultiWOZ}~\cite{budzianowski2018multiwoz} is a multi-domain dialog dataset with $10,438$ dialogs covering $7$ distinct domains and $24$ slots. We apply the agenda-based user simulator as the interaction environment for interactive evaluation.

2) \textit{SGD}~\cite{rastogi2020towards}
is published to build scalable multi-domain conversational agents. It 
has $16,142$ dialogs spanning $16$ domains and $214$ slots. We conduct a standard evaluation on SL-based methods to address the scaling concerns. We do not report interactive evaluation results since the corresponding official user simulator is unavailable.
% We present experiments on MultiWOZ~\cite{budzianowski2018multiwoz}, a multi-domain dialog dataset with $7$ distinct domains
% % \footnote{Attraction, Hospital, Police, Hotel, Restaurant, Taxi, Train}, 
% and $10,438$ dialogs.
% For standard evaluation, we  
% Each dialog trajectory is decomposed into a set of ({\em state, action, next state}) tuples used for training.
% In total, we have $56,700$ tuples in the training set, $7,300$ each for the validation set and the test set.
% % \cref{tab: data_act_count} shows the count of samples with multiple act annotations, which amounts to $68.2\%$ of the dataset.
% We apply the agenda-based user simulator~\cite{schatzmann2007agenda} as the interaction environment for interactive evaluation.
% The simulator initializes a user goal when the dialog starts and uses heuristics to provide the agent with a simulated 
% % dialog act level 
% user response at each turn.

% \footnote{Codes currently available at \url{https://www.dropbox.com/sh/7a09vm74ju9ta4o/AAA1ievLHv5INWrd2x3Wnre1a?dl=0}}.
% We use Adam as the optimization algorithm, and apply ray.Tune~\cite{liaw2018tune} to search and decide all the hyper-parameters. 
% The details are shown in the appendix.

% \subsubsection{Baselines}
\vskip 6pt
\noindent\textbf{Baselines}~~
Three types of baselines are explored: 1) Supervised Learning (SL), 2) Reinforcement Learning (RL), and 3) Adversarial Learning (AL). \mred{We ignore DQN-based methods because these methods do not suit the multi-label classification setting in our task formulation (see Sec.2 for details).}

(SL) \textit{DiaMultiClass}~\cite{li2020rethinking} is a  FFN with Sigmoid as the last activation function.

(SL) \textit{DiaSeq}~\cite{li2020rethinking} encodes the multi-hot dialog state vector with a 2-layer FFN and sequentially decodes the macro-action with an RNN.

(SL) \textit{DiaMultiDense}~\cite{li2020rethinking} performs binary classification for each specific action with the same network described in Section 3.7.

(SL) \textit{gCAS}~\cite{shu2019modeling} is a novel DiaSeq variant that uses tangled RNNs to predicts the macro-action by an auto-regressive generation of intent and slots. 

(RL) \textit{GP-MBCM}~\cite{gavsic2015policy} is a Gaussian process based Multi-domain Bayesian Committe Machine that decomposes the policy into domain-specific policies.

(RL) \textit{ACER}~\cite{wang2016sample} is the Actor-Critic RL policy with Experience Replay. ACER is sample efficient and is well adapted for large discrete action spaces.

(RL) \textit{PPO}~\cite{schulman2017proximal} denotes Proximal Policy Optimization, a simple and stable RL algorithm that applies a constant clipping mechanism.

(AL) \textit{ALDM}~\cite{liu2018adversarial} is the Adversarial Learning Dialog Model. ALDM estimates a reward at the end of the session
% with a Bi-LSTM based discrimnator
to optimize the dialog policy.

(AL) \textit{GDPL}~\cite{takanobu2019guided}
% is an adversarial inverse RL-based method that 
equips PPO with a ``rewarder'' for joint reward estimation and policy optimization. GDPL is regarded as the SOTA method.

(AL) \textit{DiaAdv}~\cite{li2020rethinking} further improves the DiaMultiDense model with adversarial fine-tuning.

\begin{table}[tp]
\centering
\small
\resizebox{.48\textwidth}{!}{
\begin{tabular}{l|ccccc}
\hline
\hline
& \multicolumn{5}{c}{\textbf{MultiWOZ}}\Tstrut\\
\hline
\textbf{Agent} & \textbf{Turn} & \textbf{Match} & \textbf{Rec}  & \textbf{F1}  & \textbf{Success} \Tstrut\\
\hline
DiaMultiClass\Tstrut & 11.46 \tiny{$\pm 0.56$} & 0.68 \tiny{$\pm 3.9\%$} & 0.81 \tiny{$\pm 3.2\%$} & 0.81 \tiny{$\pm 2.1\%$} & 67.3 \tiny{$\pm 3.69$} \\
\quad+ sample & 9.23 \tiny{$\pm 0.2$} & 0.82 \tiny{$\pm 1.1\%$} & 0.90 \tiny{$\pm 1.8\%$} & 0.77 \tiny{$\pm 1.2\%$} & 81.4 \tiny{$\pm 1.78$} \\
DiaSeq (beam) & 9.06 \tiny{$\pm0.67$} & 0.81 \tiny{$\pm 0.4\%$} & 0.9 \tiny{$\pm 1.2\%$} & 0.86 \tiny{$\pm 0.9\%$} & 81.4 \tiny{$\pm 0.16$} \\
\quad greedy & 10.35 \tiny{$\pm 0.04$} & 0.68 \tiny{$\pm 1.5\%$} & 0.80 \tiny{$\pm 0.5\%$} & 0.77 \tiny{$\pm 0.5\%$} & 67.7 \tiny{$\pm 1.02$} \\
\quad+ sample & 8.82 \tiny{$\pm 0.1$} & 0.86 \tiny{$\pm 0.6\%$} & 0.93 \tiny{$\pm 0.4\%$} & 0.81 \tiny{$\pm 0.5\%$} & 86.9 \tiny{$\pm 0.49$} \\
DiaMultiDense & 9.66 \tiny{$\pm 0.15$} & 0.85 \tiny{$\pm 0.6\%$} & 0.94 \tiny{$\pm 0.4\%$} & 0.87 \tiny{$\pm 0.6\%$} & 86.3 \tiny{$\pm 0.64$} \\
\quad- sample & 12.75 \tiny{$\pm 0.77$} & 0.61 \tiny{$\pm 6\%$} & 0.72 \tiny{$\pm 5.4\%$} & 0.80 \tiny{$\pm 2.3\%$} & 58.4 \tiny{$\pm 6.05$} \\
gCAS & 11.69 \tiny{$\pm 0.53$} & 0.56 \tiny{$\pm 1.4\%$} & 0.72  \tiny{$\pm 0.4\%$} & 0.76 \tiny{$\pm 1.4\%$} & 58.8 \tiny{$\pm 2.82$} \\
\hline
GP-MBCM\footnotemark[5]\Tstrut & \textbf{2.99} & 0.44 & -  & 0.19 & 28.9 \\
ACER\footnotemark[5] & 10.49 & 0.62 & -  & 0.78 & 50.8 \\
PPO \footnotemark[5] & 15.56 & 0.60 & 0.72 & 0.77 & 57.4 \\
\hline
ALDM\footnotemark[5]\Tstrut & 12.47 & 0.69 & - & 0.81 & 61.2 \\
GDPL & 7.54 \tiny{$\pm 0.43$} & 0.84 \tiny{$\pm 0.9\%$} & 0.89 \tiny{$\pm 2.2\%$} & \textbf{0.88 \tiny{$\pm 1.2\%$}} & 83.2 \tiny{$\pm 1.48$} \\
% \quad+ sample & 7.28 & 0.76 & 0.82 & 0.74 & 74.1 \\
DiaAdv & 8.90 \tiny{$\pm 0.18$}  & 0.87 \tiny{$\pm 0.9\%$} & 0.94 \tiny{$\pm 0.75\%$} & 0.85 \tiny{$\pm 0.58\%$} & 87.6 \tiny{$\pm 0.9$} \\
\quad- sample & 11.9 \tiny{$\pm 0.88$} & 0.62 \tiny{$\pm 5.9\%$} & 0.73 \tiny{$\pm 4.6\%$} & 0.80 \tiny{$\pm 2.1\%$} & 61.7 \tiny{$\pm 5.59$} \\
\hline
\textbf{PEDP}\Tstrut  & 8.69 \tiny{$\pm 0.15$} & \textbf{0.88} \tiny{$\pm 1.3\%$} & \textbf{0.97  \tiny{$\pm 0.4\%$}} & 0.87 \tiny{$\pm 1.1\%$}& \textbf{90.6} \tiny{$\pm 0.68$} \\
\quad- planning & 9.66 \tiny{$\pm 0.15$} & 0.85 \tiny{$\pm 0.6\%$} & 0.94 \tiny{$\pm 0.4\%$} & 0.87 \tiny{$\pm 0.6\%$} & 86.3 \tiny{$\pm 0.64$} \\
\quad- ensemble & 9.25 \tiny{$\pm 0.43$} & 0.88 \tiny{$\pm 1.97\%$} & 0.96 \tiny{$\pm 0.8\%$} & 0.85 \tiny{$\pm 2.5\%$} & 89.1 \tiny{$\pm 1.74$} \\
\quad- sample & 8.85 \tiny{$\pm 0.22$} & 0.82 \tiny{$\pm 2.5\%$} & 0.93 \tiny{$\pm 1.4\%$} & 0.86  \tiny{$\pm 1.6\%$} & 83.4 \tiny{$\pm 1.01$} \\
% \quad- sample & 12.61 & 0.62 & 0.77 & 0.81 & 63.9 \\
\hline
% {\em Human}\footnotemark[4] & \textit{7.37} & \textit{0.95} & \textit{0.86} & \textit{0.67} & \textit{75.0} \\
% \hline
\end{tabular}
}
\caption{Interactive evaluation results. We simulate 1,000 dialogs per run and report the mean and standard deviation over $5$ runs.}
\label{tab: robust}
\end{table}
\footnotetext[5]{Results reused from~\cite{li2020rethinking}}
% \footnotetext[4]{Results reused from~\cite{takanobu2019guided}}

\subsection{Model Evaluation}
\textbf{Interactive Evaluation}~~
% We compare the performance of different models in three aspects:
% 1) {\em Robust Evaluation}, which reports the quantitative indicators of task completion,
% 2) {\em Standard Evaluation}, which reports 
% \subsubsection{\mred{Interactive Evaluation}}
We report the performance of each approach that interacts with the user simulator in \cref{tab: robust}.

We observe that our PEDP framework achieves the highest performance in the task success by $3\%$ compared to the second-best model DiaAdv and by $7.4\%$ compared to the SOTA GDPL method on account of the substantial improvement in match rate and inform recall. 
% Compared to the DiaMultiDense model with Gumbel-Sigmoid sampling (i.e., PEDP without planning), PEDP also has significant improvement in terms of all robustness metrics.
This demonstrates that
% the proposed {em single-action planning module} in our 
PEDP can effectively incorporate more task-relevant information to enrich the response and lead to task success.

% PEDP also has significant improvement compared to the DiaMultiDense model that applies the same decoder and action sampling mechanism as our method.
% The task success is improved by $4.3\%$.
% Note that fine-tuning the DiaMultiDense with adversarial learning (DiaAdv) only brings a marginal improvement of the task success by $1.3\%$. 
% This indicates the effectiveness of the single-action dialog planning module in our PEDP framework.

% Compared to the DiaMultiDense model that shares the same decoder and action sampling mechanism with our PEDP framework and the DiaAdv model that fine-tunes the DiaMultiDense with adversarial learning, our method also has significant improvement in terms of all evaluation metrics.
% The task success is improved by $4.3\%$ and $3\%$, respectively.
% This indicates the effectiveness of the single-action dialog planning module in our PEDP framework. 

% The PEDP framework also beats human in task completion.
% We observe that human share superior match rate and have rather lower F1 and higher recall for inform rate, which is also the case in our PEDP method.
% A possible reason is that human almost manages to make a reservation in each session, but may forget to ask for all the required information, as reported in~\cite{takanobu2019guided}.

We also observe that PEDP performs moderately in terms of average turn.
A possible reason is that the model fails to generate denser dialog action combinations.
As shown in \cref{tab: robust}, PEDP reaches a comparable F1 score and a higher recall, indicating a lower precision score and implying the incorporation of redundant information that does not contribute to task completion.
We report how this affects user experience in the human evaluation.
% SOTA adversarial training method DiaAdv and GDPL, PEDP achieves $3\%$ and $8.3\%$ improvements, respectively.
% by $3\%$ and $8.3\%$ compared to the SOTA adversaral training method DiaAdv and GDPL, respectively.
% On account of the substantial improvement in inform recall and match rate, our PEDP model achieves the highest performance in the task success of over $90\%$ by $3\%$ and $8.3\%$ compared to the SOTA adversaral training method DiaAdv and GDPL, respectively.
% Compared to the vanilla DiaMultiClass model with Gumbel-Sigmoid sampling (i.e., PEDP without planning), PEDP also has significant improvement in terms of all robustness metrics.
% The success score is improved by $9.2\%$.
% This demonstrates the effectiveness of the single-action planning module.
% Since the single-action planning module implicitely discovers and models sequential orders between the atomic actions, it can always encourage the dialog policy to incorporate more related atomic actions for macro-action prediction.
% However, with a high inform recall score, the PEDP model have a rather moderate inform F1 score, which means a lower precision.
% This demonstrates the incorporation of unrelated atomic actions, which we assume is caused by highly diversed single-action planning and can be controled with the temperature factor.
% We account the observation that PEDP do not beat GDPL in terms of average turns for the same reason since the redundant actions tends to trigger extra reponses.

\begin{table}[tp]
\centering
\small
\resizebox{.48\textwidth}{!}{
\begin{tabular}{l|ccc|ccc}
\hline
\hline
\vspace{1pt}
% \vskip 1pt
& \multicolumn{3}{c|}{\textbf{MultiWOZ}} & \multicolumn{3}{c}{\textbf{SGD} (scaling)} \Tstrut\\
\hline
\textbf{Agent} & \textbf{F1}\% & \textbf{Precision}\% & \textbf{Recall}\% & \textbf{F1}\% & \textbf{Precision}\% & \textbf{Recall}\% \Tstrut\\
\hline
DiaMultiClass\Tstrut & 39.41 \tiny{$\pm 1.08$} & 54.59 \tiny{$\pm 1.71$} & 34.32 \tiny{$\pm 1.32$} & 58.09 \tiny{$\pm 0.63$} & 81.29 \tiny{$\pm 1.13$} & 46.29 \tiny{$\pm 0.57$} \\
\quad+ sample & 38.91 \tiny{$\pm 0.74$} & 47.28 \tiny{$\pm 0.68$} & 37.56 \tiny{$\pm 1.08$}  & 58.03 \tiny{$\pm 0.64$} & 81.48 \tiny{$\pm 0.18$} & 46.14 \tiny{$\pm 0.80$}\\
DiaSeq (beam) & 44.64 \tiny{$\pm 2.08$} & 51.91 \tiny{$\pm 0.99$} & 43.66 \tiny{$\pm 2.27$} & 63.13 \tiny{$\pm 0.18$} & 86.04 \tiny{$\pm 0.5$} & 50.83 \tiny{$\pm 0.30$}\\
\quad greedy & 48.34 \tiny{$\pm 0.45$} & 54.71 \tiny{$\pm 0.21$} & 48.84 \tiny{$\pm 0.84$} & 63.21 \tiny{$\pm 0.35$} & 86.31 \tiny{$\pm 0.7$} & 50.85 \tiny{$\pm 0.40$}\\
\quad + sample & 37.82 \tiny{$\pm 0.45$} & 43.02 \tiny{$\pm 0.48$} & 38.91 \tiny{$\pm 0.64$} & 62.64 \tiny{$\pm 1.03$} & 85.54 \tiny{$\pm 1.62$} & 50.40 \tiny{$\pm 0.76$} \\
DiaMultiDense & 35.92 \tiny{$\pm 0.54$} & 51.93 \tiny{$\pm 0.33$}  & 30.10 \tiny{$\pm 0.69$} & 57.85 \tiny{$\pm 0.68$} & 80.64 \tiny{$\pm 0.43$}  & 46.21 \tiny{$\pm 0.89$}  \\
\quad- sample & 34.35 \tiny{$\pm 0.62$} & 52.14 \tiny{$\pm 0.19$} & 27.74 \tiny{$\pm 0.74$} & 56.69 \tiny{$\pm 0.62$} & 79.54 \tiny{$\pm 0.88$} & 45.19 \tiny{$\pm 0.75$}\\
gCAS\Bstrut & 50.01  \tiny{$\pm 0.62$} & 55.56  \tiny{$\pm 0.59$}& 51.21  \tiny{$\pm 1.74$}  & 76.37  \tiny{$\pm 1.60$} & 77.70  \tiny{$\pm 1.46$}& 79.99  \tiny{$\pm 1.03$}\\
\hline
GDPL & 31.89\Tstrut \tiny{$\pm 0.96$} & 50.14 \tiny{$\pm 0.79$} & 24.99 \tiny{$\pm 1.14$} & - & - & - \\
\quad+ sample & 34.60 \tiny{$\pm 0.47$} & 45.01 \tiny{$\pm 0.24$} & 31.54 \tiny{$\pm 0.80$} & - & - & -\\
DiaAdv & 40.97 \tiny{$\pm 0.95$}  & 53.44 \tiny{$\pm 0.50$} & 36.84 \tiny{$\pm 1.30$} & - & - & - \\
\quad- sample\Bstrut & 41.71 \tiny{$\pm 0.47$} & 56.46 \tiny{$\pm 0.45$} & 36.28 \tiny{$\pm 1.48$} & - & - & -\\
\hline
\textbf{PEDP}\Tstrut & 64.63  \tiny{$\pm 0.16$} & 77.03 \tiny{$\pm 1.39$} & 61.77 \tiny{$\pm 1.01$} & 84.12  \tiny{$\pm 0.38$} & 91.66 \tiny{$\pm 0.52$} & 81.19 \tiny{$\pm 0.4$}\\
\quad- planning & 35.92 \tiny{$\pm 0.54$} & 51.93 \tiny{$\pm 0.33$}  & 30.10 \tiny{$\pm 0.69$} & 57.85 \tiny{$\pm 0.68$} & 80.64 \tiny{$\pm 0.43$}  & 46.21 \tiny{$\pm 0.89$}  \\
\quad- ensemble &  64.34 \tiny{$\pm 0.29$} & 77.63 \tiny{$\pm 2.04$} & 60.85 \tiny{$\pm 1.54$}  &  83.31 \tiny{$\pm 0.55$} & 91.66 \tiny{$\pm 0.78$} & 80.10 \tiny{$\pm 0.55$}\\
\quad- sample\Bstrut & \textbf{66.95 \tiny{$\pm 0.45$}} & \textbf{78.11 \tiny{$\pm 3.03$}} & \textbf{65.02 \tiny{$\pm 1.22$}} & \textbf{84.74 \tiny{$\pm 0.55$}} & \textbf{92.07 \tiny{$\pm 0.97$}} & \textbf{81.30  \tiny{$\pm 0.82$}}\\
% \textbf{PEDP} & 34.23 & 48.91 & 29.01 \\
% \quad- planning & 35.92 \tiny{$\pm 0.54$} & 51.93 \tiny{$\pm 0.33$}  & 30.10 \tiny{$\pm 0.69$}  \\
% \quad- sample & 35.64 & 54.87 & 28.29 \\
\hline
\end{tabular}
}
\caption{Standard evaluation results. We report the mean and standard deviation over $5$ runs.}
%  We do not report results for AL-based models on the SGD dataset because the corresponding official user simulator is not available.
\label{tab: standard}
\end{table}

\vskip 6pt
\noindent\textbf{Standard Evaluation}~~
We report the standard performance of the state-of-the-art GDPL and DiaAdv and all the SL-based methods in \cref{tab: standard}.

\mred{We observe that our PEDP framework outperforms the others in terms of all evaluation metrics. For the F1 score, PEDP achieves $8.37\sim32.74\%$ improvements on account of the substantial improvement in precision ($6.03\sim26.89\%$) and recall ($1.31\sim36.78\%$) over the baselines.}

\mred{We also observe that the standard and interactive performance are negatively correlated for the baseline methods (see \cref{tab: robust}). We attribute this to the fact that standard and interactive objectives are fundamentally in conflict. While multiple action combinations can be considered appropriate during dialog interaction, they may not match the ground truth in the test samples. Unlike existing methods, PEDP achieves the best of both worlds, demonstrating our method's effectiveness on MADPL.}

\vskip 6pt
\noindent\textbf{Ablation Study}~~
\mred{We conduct ablation experiments to investigate the effects of two components in our model (Single-action Dialog Planning and Ensemble Prediction) and study how stochastic factors affect model performance (action sampling). We merge the results in Table~\ref{tab: robust} and \ref{tab: standard}.}

% \mred{To verify the effectiveness of the single-action dialog planning module, we consider a no-planning version of PEDP (i.e., ``- planning''), which equals the DiaMultiDense model. To evaluate the importance of the ensemble prediction, we report the results with a single planned path (i.e., ``- ensemble''). To study how stochastic factors affect model performance, we build model variants that generate the macro-action with (without) probability sampling, i.e., ``+ (-) sample'', for the SL-based methods, GDPL, and DiaAdv.}
% The results show that: 1) the single-action dialog planning module can effectively incorporate relevant dialog actions and boost performance by improving match rate and recall while improving/preserving precision. 2) ensemble prediction can effectively reduce the impact of low-quality paths and lead to a lower standard deviation\footnote{Please see Appendix D for hyper-parameter sensitivity experiments on number of paths $K$.}. 3) Sampling usually leads to significant improvement of match rate and inform recall but reduces the inform precision, indicating the incorporation of redundant information in the response. For GDPL, sampling harms the performance. A possible reason is that GDPL, which learns from the user simulator for evaluation, has adapted to the simulated dialog scenarios. 
% Sampling instead may lead to sub-optimal decisions.

\mred{The results show that: 1) The single-action dialog planning module contributes more significantly for performance improvement while the ensemble prediction module effectively stabilizes the model and leads to a lower standard deviation, and please see Appendix C for hyper-parameter sensitivity experiments for the number of paths $K$. 2) Sampling usually leads to a performance gain of match rate and inform recall but reduces the inform precision, indicating the incorporation of redundant information in the response. For GDPL, sampling harms the performance. A possible reason is that GDPL, which learns from the user simulator for evaluation, has adapted to the simulated dialog scenarios. 
Sampling instead may lead to sub-optimal decisions.}

\begin{table}[tp]
\centering
\small
% \resizebox{.5\textwidth}{!}{
\begin{tabular}{l|ccc|c}
\hline
\hline
\textbf{Dialog pair} & \textbf{Win} & \textbf{Lose} & \textbf{Tie} & $\mathbf{\alpha}$ \\
\hline
PEDP vs. DiaSeq & 41.7 & 31.3 & 27.0 & 0.820\\
\hline
PEDP vs. DiaAdv & 36.5 & 27.6 & 35.9 & 0.856 \\
\hline
PEDP vs. GDPL & 32.6 & 26.5 & 40.9 & 0.839 \\
\hline
\end{tabular}
% }
\caption{Human evaluation results. We report the mean over $9$ judges and Krippendorff's alpha ($\alpha$) that measures the inter-rater reliability. Typically, results are considered reliable if $\alpha > 0.800$.}
\label{tab: human}
\end{table}

\vskip 6pt
\noindent\textbf{Human Evaluation}~~
\mred{Automatic evaluation only measures part of the agent's performance (e.g., {\em Success} for the level of task completion). It may fail to consider other aspects for assisting real users (e.g., inappropriate reply).
Thus, we conduct a human study to fully evaluate system performance.}
Our PEDP method is compared with three state-of-the-art dialog policy models: GDPL, DiaAdv, and DiaSeq
(reported to offer an elegant user experience~\cite{li2020rethinking}).
\mred{Following~\cite{takanobu2019guided,li2020rethinking}}, for each comparison pair, we randomly sample 100 user goals
% containing the constraints about specific domain slots and slot information that the user is looking for. Then, 
and present the generated dialog samples from the two agents.
The dialog actions are translated into human-readable utterances with the language generation module from ConvLab~\cite{zhu2020convlab}.
\mred{We hire nine postgraduates} as judges to select the dialogs that provide a better user experience (each sample is graded nine times rather than three in existing works~\cite{takanobu2019guided,li2020rethinking}), ``Tie'' can also be selected for the equally performed ones.
We refer the reader to Appendix D and E for the case study and the corresponding error analysis.

In \cref{tab: human}, we see that our PEDP framework outperforms the baseline models.
As mentioned in robust evaluation (see Section 5.4), our PEDP model shares a lower inform precision ($0.79$) than DiaSeq ($0.82$) and GDPL ($0.87$), which indicates the incorporation of redundant information.
However, we find that the performance gap does not severely harm the user experience.
A possible reason is that human users tend to focus on goal-related information 
% and easily ignore additional content 
during task-oriented dialog.
Compared to dealing with redundant information, they are more intolerant towards the failure of providing the requested information.

\section{Conclusion}
This paper presents a supervised learning-based PEDP framework, which incorporates model-based planning for conceiving what to express before deciding the current response through simulating single-action dialogs.
With multi-task learning of single-action dialog dynamics and multi-action prediction, our method fully uses the training corpus and makes reasonable and explainable decisions, thus improving performance towards unseen task-oriented human-machine dialog flows.
Extensive experiments show that our method significantly and consistently outperforms state-of-the-arts with a favorable task success rate of $90.6\%$.

\section*{Acknowledgments}
The research presented in this paper is supported in part by National Key R\&D Program of China (2021YFB1715600), National Natural Science Foundation of China (61902305, 61922067, U1736205), MoE-CMCC “Artifical Intelligence” Project (MCM20190701),  Shenzhen Basic Research Grant (JCYJ20170816100819428), Natural Science Basic Research Plan in Shaanxi Province of China (2019JM-159), and Natural Science Basic Research Plan in Zhejiang Province of China (LGG18F020016).

\bibliographystyle{named}
\bibliography{ijcai22}
% \appendix
% \input{Main/main}
\end{document}